\pdfoutput=1

\documentclass[11pt]{article}

\usepackage{authblk}
\usepackage{emnlp2021}
\usepackage[ruled,vlined]{algorithm2e}
\usepackage{amssymb}
\usepackage{bm}
\usepackage{booktabs}
\usepackage{times}
\usepackage{mathtools}
\usepackage{amsmath}
\usepackage{latexsym}
\usepackage{verbatim}
\usepackage{graphicx}
\usepackage{adjustbox}
\usepackage{caption}
\usepackage{subfigure}
\usepackage{multicol}
\usepackage[T1]{fontenc}

\usepackage[utf8]{inputenc}

\usepackage{microtype}

\makeatletter
\renewcommand\AB@affilsepx{\quad \protect\Affilfont}
\makeatother

\title{Self-training with Few-shot Rationalization: \\ Teacher Explanations Aid Student in Few-shot NLU}

\author[$\dagger$]{\bf{Meghana Moorthy Bhat}}
\author[$\ddagger$]{\bf{Alessandro Sordoni}}
\author[$\ddagger$]{\bf{Subhabrata Mukherjee}}

\affil[$\dagger$]{{The Ohio State University}} \affil[$\ddagger$]{{Microsoft Research}} 
\affil[ ]{\authorcr bhat.89@osu.edu, \{alsordon, submukhe\}@microsoft.com}

\begin{document}
\maketitle
\begin{abstract}
While pre-trained language models have obtained state-of-the-art performance for several natural language understanding tasks, they are quite opaque in terms of their decision-making process. While some recent works focus on rationalizing neural predictions by highlighting salient concepts in text as justifications or rationales, they rely on thousands of labeled training examples for both task labels as well as annotated rationales for every instance. Such extensive large-scale annotations are infeasible to obtain for many tasks. To this end, we develop a multi-task teacher-student framework based on self-training language models with limited task-specific labels and rationales, and judicious sample selection to learn from informative pseudo-labeled examples\footnote{Code available at \url{https://aka.ms/RationaleST}}. We study several characteristics of what constitutes a good rationale and demonstrate that the neural model performance can be significantly improved by making it aware of its rationalized predictions particularly in low-resource settings. Extensive experiments in several benchmark datasets demonstrate the effectiveness of our approach.

\end{abstract}

\section{Introduction}

Recent success in several natural language understanding tasks can be attributed to training large-scale and complex neural network models. While these models work very well for specific tasks, they offer limited insights into their inner working and are often used as black-box predictors. To address these shortcomings, recent works~\cite{deyoung-etal-2020-eraser, paranjape-etal-2020-information, yu-etal-2019-rethinking} have focused on designing interpretable NLP systems that can explain the model's predictions. A typical approach to study this decision-making process has been via annotating {\em rationales} as a short and sufficient part of the input text leading to the specific prediction that can be also used as auxiliary supervision for training. Appropriate use of such rationales can improve the downstream task performance as the model learns to focus on the task-relevant parts of the input~\cite{ pruthi2020evaluating}.

However, a significant resource challenge is to obtain large-scale annotated rationales to train these models as explored in fully supervised setting in recent work~\cite{deyoung-etal-2020-eraser}. This requires models to have access to both instance-level task labels as well as token-level binary rationale labels depicting whether a token should be included in the rationale or not. Such extensive annotations are infeasible to obtain for many tasks, hence devising models that can effectively exploit a limited number of annotated rationales is utterly important.
Therefore, our objective is two-fold: (1) improve downstream task performance and (2) improve rationale extraction -- with few labeled examples for the downstream task and corresponding rationales.

Recent works~\cite{NEURIPS2020_f23d125d, DBLP:journals/corr/abs-2010-03680, DBLP:journals/corr/abs-1911-04252} on few-shot learning have explored \emph{self-training} as a mechanism to train neural network models with limited labeled data. These methods usually train a teacher and then a student model to imitate the teacher in turn. They usually assume access to a set of unlabeled instances and use stochastic regularization techniques such as dropout and data augmentation obtained from pseudo-labeled examples. In this work, we leverage self-training as a mechanism to train neural network models with self-generated rationales and task labels over unlabeled data. Since pseudo-labeled rationales from the teacher model can be noisy, we show that judicious sample selection to upweight informative examples and downweight noisy ones is beneficial. Furthermore, we predict task and rationale labels in a multi-task learning (MTL) setup, where we share parameters between the task objective and the rationale prediction objective. We show that the MTL setup for joint learning is more effective than the decoupled learning, which consists of first extracting rationales and then using them for classification, as explored in some of the prior works.


Given the paucity of rationale labels, a critical part in the MTL setup is to understand what constitutes a good rationale. We build over insights from prior work~\cite{yu-etal-2019-rethinking,lei-etal-2016-rationalizing} focusing on low-resource settings with access to limited labels via multi-task self-training. To this end, we explore several characteristics of a good rationale in terms of (i) {\em sufficiency} such that the extracted rationale is adequate for the model to make its decision; (ii) {\em completeness} such that the model is less confident on its predictions if it ignores the rationale text; and (iii) {\em coherency} such that the model extracts phrases as rationales rather than isolated disconnected words. In practice, we enforce (i) by matching predictions of the student model with the rationale as input and the teacher model with the full input; (ii) by maximizing entropy in the student predictive distribution when it sees the complementary of the rationale as input; and (iii) by recurring to additional regularization methods.
We show that our multi-task joint optimization captures all of the above salient aspects for rationale extraction while improving the downstream task performance. In summary, our contributions are:

\noindent{\bf(a)} We develop a multi-task self-training framework to train neural models with limited labels along with extracting rationales as justifications for its predictions. Furthermore, we show the impact of judicious sample selection for sample- and token-level re-weighting to learn from informative pseudo-labeled examples during self-training.

\noindent{\bf(b)} We build over prior work on rationale extraction to encode desired rationale characteristics by judiciously designed loss functions in our multi-task self-training algorithm.

\noindent{\bf(c)} Extensive experiments on five datasets from the ERASER benchmark~\cite{deyoung-etal-2020-eraser} demonstrate the effectiveness of our approach. Ablation experiments demonstrate the impact of different components of our framework.

\section{Related Work}
\noindent{\bf Rationale Extraction} Prior works~\cite{lei-etal-2016-rationalizing,yu-etal-2019-rethinking} on rationale extraction explore encoder-generator based models with two components for extracting rationales and then using them to make a prediction. Alternately, \citet{jain-etal-2020-learning} propose decoupled architectures for extractor (using attention weights) and predictor. 
Following these works, \citet{deyoung-etal-2020-eraser} develop the ERASER benchmark that contains human annotated rationales in extractive format and provide BERT-to-BERT baselines for the tasks. 
\citet{paranjape-etal-2020-information} propose a weakly supervised model with user controlled sparsity threshold for rationale extraction and predictions based on the extracted rationale. Similarly, \citet{pruthi-etal-2020-weakly} propose a semi-supervised BERT-CRF architecture with few gold annotations and abundant task labels. 
In contrast to all these prior work requiring thousands of annotations for either rationales or the task labels, our framework is geared for low-resource settings with access to very few labels for both the tasks.
We incorporate insights from prior work in rationale extraction via judiciously designed loss functions in our multi-task self-training framework.

\paragraph{Self-training} Self-training \cite{yarowsky-1995-unsupervised, 10.1145/354756.354805, Lee2013PseudoLabelT} trains a base (teacher) model on limited labeled data and applies them to unlabeled data to generate pseudo-labels. The generated pseudo-labels are used for training the student model in an iterative fashion. Self-training has demonstrated state-of-the-art performance in several tasks including text classification~\cite{NEURIPS2020_f23d125d, DBLP:journals/corr/abs-2010-03680} and image classification~\cite{DBLP:journals/corr/abs-1911-04252, zoph2020rethinking}. We leverage self-training with re-weighting noisy pseudo-labels for both task and rationale extraction in a multi-task learning framework while encoding the desired characteristics of a good rationale. 


\section{Rationale Extraction with Few Labels}
\begin{figure*}
\centering
\includegraphics[width=1\textwidth]{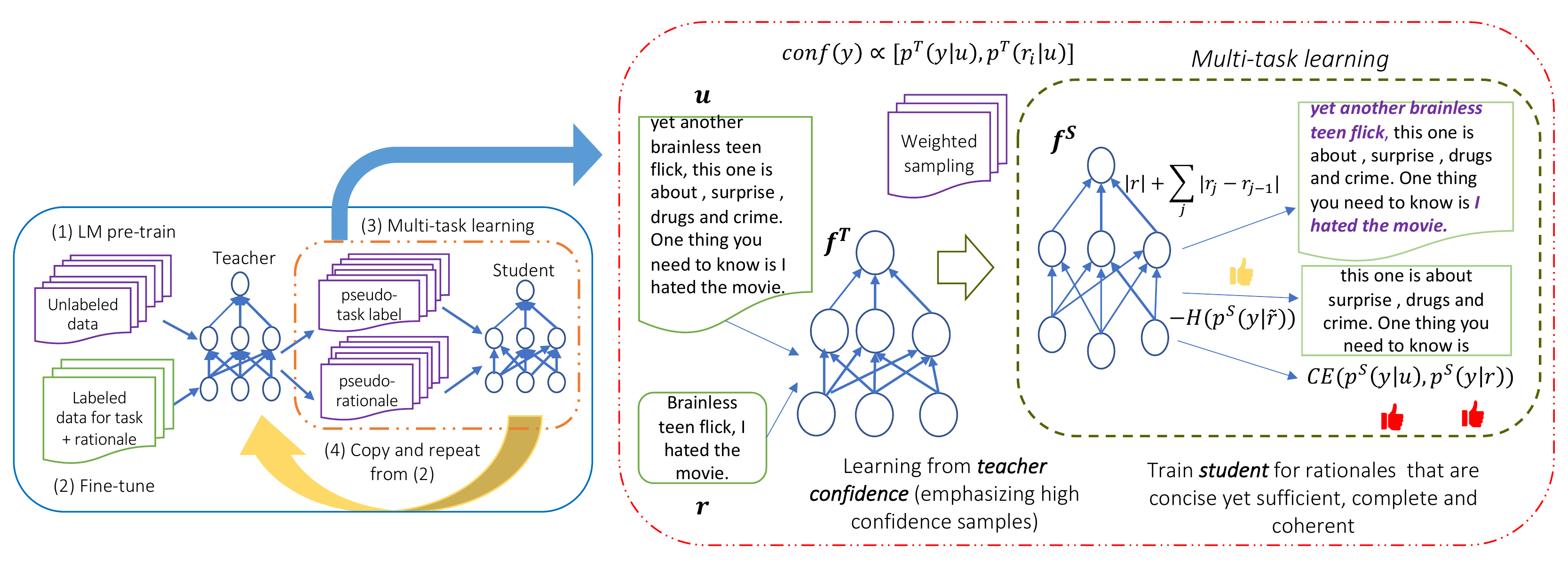}
\vspace*{-2em}
\caption{Self-training framework}
\vspace*{-1em}
\label{fig:framework_outline}
\end{figure*}

\subsection{Problem Statement}
Let $\bm{x}_1, \ldots, \bm{x}_n$ be a set of $n$ documents with corresponding associated task labels $y_1, \ldots, y_n$, where each $\bm{x}_i=\{x_{ij}\}$ is a sequence of tokens. We consider each document to be associated with a ground-truth rationale sequence $\bm{r}_i=\{r_{ij}\}$, where $r_{i, j} = 1$ if the $j^{th}$ token in document $\bm{x}_i$ is part of the rationale, and 0 otherwise. We consider a {\em low-resource} setup with {\em very few} documents labeled with both the task labels (instance-level) and the rationale labels (token-level) for each task, and additional unlabeled data.

Let us denote $\mathcal{D}_l = \{(\bm{x}_i, y_i, \bm{r}_i)\}$ as the joint task and rationale labeled training set. We also assume to have access to a set of unlabeled documents $\mathcal{D}_u = \{\bm{u}_1, \ldots, \bm{u}_m \}$ for which neither rationales nor task labels are available and $|\mathcal{D}_u| \gg |\mathcal{D}_l|$. 
Our goal is to learn a model from the few task and token-level rationale labels and additional unlabeled documents to improve its performance for the downstream tasks.

\begin{algorithm}[t]
\small
Initialize $p^T, p^S$ randomly\\
\While{Not converged}{
$p^T = \min_{p^T} \mathcal{L}_{l}(p^T)$ \emph{\# Eq. (\ref{eq:loss_lab})} \\
$p^S = \min_{p^S} \mathcal{L}_{u}(p^S)$ \emph{\# Eq. (\ref{eq:joint})} \\
Copy $p^S$ into $p^T$.
}
return $p^S$
\caption{Our self-training algorithm.}
\end{algorithm}

\subsection{Self-training}
We leverage self-training as the backbone of our framework. The algorithm is composed of two phases that are executed iteratively until convergence. In the first phase, we perform multi-task learning of a teacher model on the few-shot labeled set $\mathcal{D}_l$ by jointly learning to predict instance-level task and token-level rationale labels. Optimizing these losses leads to estimating the parameters of a {\em teacher} model $p^T$.

In the second phase, we leverage the teacher model $p^T$ to infer pseudo-labels for the unlabeled set $\mathcal{D}_u$ and train a {\em student} model $p^S$ to mimic the teacher's predictions for both the task and associated rationale. Finally, the teacher model is updated with the student model's parameters and above steps are repeated until convergence.  

Due to the noisy nature of pseudo-labels, the above self-training process may result in gradual drifts~\cite{zhang2016understanding}. To address this, we train the student model to explicitly account for the teacher's confidence on the generated pseudo-labels with a special weighting scheme. Furthermore, we explore several characteristics of a good rationale and enrich the above framework with additional auxiliary losses.


\subsubsection{Multi-Task Teacher} 

In the first phase, we leverage the small amount of labeled data $\mathcal{D}_l$ to train the teacher model $p^T$ to jointly predict the task labels and the rationale labels. To this end, we leverage a {\em shared} BERT encoder $h^T$ with two separate softmax classification layers for the two tasks. We denote $p^T(y | \bm{x}) = \text{softmax}(h^T(\bm{x}); \theta^T_{t})$ and $p^T(r_j | \bm{x}) =\text{softmax}(h^T(\bm{x})_j; \theta^T_{r})$ to be the corresponding task and rationale predictions of the teacher model given an instance $\bm{x}$. $h^T(\bm{x})_j$ is the BERT hidden state representation corresponding to the $j^{th}$ token and $\theta^T_{t}$,$\theta^T_{r}$ are the task-specific head parameters. For brevity, we will omit parameter specification in what follows and denote $p^T$ as the shared BERT and task-specific parameters.

We jointly optimize the following losses with respect to (the parameters of) $p^T$:

\begin{align}
\small
\begin{split}
\mathcal{L}&_{l}(p^T) = \mathbb{E}_{\mathcal{D}_l} \left[- \underbracket{\log p^T(y | \bm{x})}_{\text{task prediction}} - \underbracket{\sum_j \log p^T(r_j | \bm{x})}_{\text{rationale prediction}} \right],
\end{split}
\label{eq:loss_lab}
\end{align}

where $y$ is the ground-truth task label for input $\bm{x}$ and $\bm{r}^T_{j}$ is the ground-truth rationale label for the $j^{th}$ token in input $\bm{x}$.

In contrast to prior work~\cite{deyoung-etal-2020-eraser} that leverages two decoupled BERT models in a stage-wise fashion to first extract the rationales, and then use those rationales for task label prediction -- our framework learns a single model to predict them jointly in a multi-task learning setup. This allows the model to capture richer interactions between the two tasks.

There are a few design choices for optimizing the teacher in Eq.~\ref{eq:loss_lab}. For instance, we can optimize the teacher parameters with few-shot labeled data in {\em each} self-training iteration after the student becomes the new teacher; or {\em only once} at the beginning of self-training to initialize a good teacher. We observe that executing this phase in each self-training loop is more effective as it diminishes drifting from the ground-truth data distribution.

\subsubsection{Multi-Task Student}
In the second phase, we self-train a student model $p^S$ on the teacher-generated pseudo-labels with a pseudo-label task loss and rationale loss. In contrast to the teacher model operating on the few labeled examples $\mathcal{D}_l$, the student model operates on unlabeled data $\mathcal{D}_u$. Additionally, the student model has the same architecture as the teacher model with a shared encoder $h^S$ and task-specific classification parameters $p^S(y | \bm{x}) = \text{softmax}(h^S(\bm{x}); \theta^S_{t})$ and $p^S(r | \bm{x}) =\text{softmax}(h^S(\bm{x}); \theta^S_{r})$. The pseudo-labeled multi-task loss is formulated as:
\begin{align}
\begin{split}
\mathcal{L}_{u}&(p^S) = \mathbb{E}_{\bm{u} \sim \mathcal{D}_u, \bm{r}^T \sim p^T(\bm{r} | \bm{u}), y^T \sim p^T(y | \bm{u})} \left. \biggl[ \right. \\
& \left. -\underbracket{\log p^S(y^T | \bm{u})}_{\text{task pseudo prediction}} - \underbracket{\sum_j \log p^S(r^T_j | \bm{u})}_{\text{rationale pseudo prediction}} \right. \biggr]
\end{split}
\label{eq:pseudo_loss}
\end{align}
where $y^T$ is the teacher-generated task pseudo-label for input $\bm{u}$ and $\bm{r}^T_{j}$ is the teacher-generated rationale pseudo-label for the $j^{th}$ token in input $\bm{u}$.

\subsubsection{Student-Teacher Update} At the end of every self-training iteration, we transfer the knowledge acquired by the student back into the teacher model by setting $h^T=h^S, \theta^T_t=\theta^S_t, \theta^T_r = \theta^S_r$ and start again by fine-tuning the newly obtained teacher on ground-truth data $\mathcal{D}_l$.

\subsubsection{Re-weighting Pseudo-labeled Samples}
Instead of directly imitating the teacher's predictions as described in Eq.~\ref{eq:pseudo_loss}, we found it extremely effective to train the student model to explicitly account for the teacher's confidence for the generated pseudo-labels. This allows us to filter noisy pseudo-labels as the student model can selectively focus more on the pseudo-labeled samples that the teacher is more confident on compared to the less certain ones.
Therefore, we optimize a weighted version of the pseudo-labeled loss in Eq.~\ref{eq:pseudo_loss}:
\begin{align}
\begin{split}
\mathcal{L}&_{wu}(p^S) = \mathbb{E}_{\bm{u} \sim \mathcal{D}_u, \bm{r}^T \sim p^T(\bm{r} | \bm{u}), y^T \sim p^T(y | \bm{u})} \left. \biggl[ \right. \\
& - \underbracket{w_y^T(\bm{u})}_{\text{task weight}} \cdot\, \log p^S(y^T | \bm{u}) \, \\
& - \sum_j \underbracket{w^T_{r, j}(\bm{u})}_{\text{rationale weight}} \cdot\, \log p^S(r^T_j | \bm{u}) \left.  \right. \biggr],
\end{split}
\end{align}
where $w_y^T(\bm{u}) \propto p^T(y = y^T | \bm{u})$ and $w^T_{r, j}(\bm{u}) \propto p^T(r_{j} = r^T_{j} | \bm{u})$. The proportional sign is due to the fact that these weights are normalized across each batch when training by minibatch SGD, so the weights \emph{depend on the batch} and sum to one over the batch. Re-weighting noisy labels with different weighting schemes has been explored with meta-learning~\cite{DBLP:conf/icml/RenZYU18} and uncertain-aware self-training~\cite{NEURIPS2020_f23d125d}. 

\subsection{Rationale Characteristics}

In this section, we encode several characteristics of what constitutes a good rationale from prior work in our self-training framework via several auxiliary loss functions.

\subsubsection{Sufficiency}
A desired property of a good rationale is \emph{sufficiency}. This imposes the model predictions about the task label considering the entire input text to be similar to the predictions made by looking at only the rationale text. 
This concept can be promptly translated into an objective function by resorting to a consistency objective:
\begin{align}
\begin{split}
\mathcal{L}_{suff}(p^S) =\, \mathbb{E} \biggl[ 
- \log p^S(y = y^T | \bm{u} \odot \bm{r}^T) \biggr],
\end{split}
\label{eqn:suff_loss}
\end{align}
where the expectation is taken w.r.t $\bm{u}\sim\mathcal{D}_u$, $\bm{r}^T \sim p^T(\bm{r} | \bm{u})$, $y^T \sim p^T(y | \bm{u})$ and $\bm{u} \odot \bm{r}^T$ is the masked version of document $\bm{u}$ in which tokens that are not part of the rationale (as predicted by the teacher) are replaced with a special \texttt{[MASK]} token. Here, the teacher model looks at the full input and the student model looks only at the rationale tokens.

The sufficiency loss can be interpreted as an alternative way of integrating rationale information in the model.  Current efforts either predict the rationale first and then use it for task prediction sequentially as in BERT-to-BERT~\cite{deyoung-etal-2020-eraser}; or employ attention regularization such that the BERT attention weights are as close as possible to uniform on the rationale tokens~\cite{pruthi2020evaluating}. The first approach can be very sensitive to error propagation from the rationale generator since the task label is predicted using only the generated rationales at test-time. The second approach strictly assumes uniform attention on the rationale tokens. In contrast, our sufficiency loss makes very few assumptions on how the model should attend to the rationale tokens, and only requires the student distribution of the task labels to be similar to that of the teacher. This yields more robustness to rationale errors, given that at test-time our model can use the full input to predict the task label.

\subsubsection{Completeness}
Another desiderata of a rationale is \emph{completeness}. Completeness implies that the rationale should capture all the aspects in the input text that are predictive of the task label. We translate this concept by requiring the student model to be \emph{maximally uncertain} of the task label if it does not look at the rationale i.e. by masking out the teacher-predicted rationale tokens in the input text:
\begin{align}
\begin{split}
\mathcal{L}_{comp}(p^S) =\, & \mathbb{E} \biggl[ - H(p^S(y | \bm{u} \odot (1 - \bm{r}^T))) \biggr]
\end{split}
\end{align}
where the expectation is w.r.t. $\bm{u}\sim\mathcal{D}_u, \bm{r}^T \sim p^T(\bm{r} | \bm{u})$, $H$ is the entropy of the student predictive distribution and $\bm{u} \odot (1 - \bm{r}^T)$ is the document obtained by masking out the tokens in the rationale.

\subsubsection{Coherence Loss}
Finally, we desire the rationales to be short and composed of contiguous chunks of text rather than unigrams. To ensure this, we adopt the regularization losses introduced in~\citet{lei-etal-2016-rationalizing}. This explicitly penalizes the rationale generator for predicting long rationales and encourages rationales to span contiguous chunks of the input text.
\begin{align}
\begin{split}
\mathcal{L}_{co}(p^S) = \mathbb{E}_{\bm{u}, \bm{r}^S \sim p^S(\bm{r} | \bm{u})} \biggl[|\bm{r}^S| + \sum_j |r^S_{j} - r^S_{j - 1}| \biggr]
\label{eqn:coherence_loss}
\end{split}
\end{align}

\subsection{Training Objectives}
Our overall training objective in the teacher learning phase is simply the loss on labeled data $\mathcal{L}_{l}(p^T)$. For the student, we use a combination of the previously presented loss functions on unlabeled data:
\begin{align}
\begin{split}
\mathcal{L}_u(p^S) = \mathcal{L}_{wu} + \mathcal{L}_{suff} + \mathcal{L}_{comp} + \mathcal{L}_{co}
\label{eq:joint}
\end{split}
\end{align}

\section{Experimental Setup}

\paragraph{\bf Datasets} 
We evaluate our framework on five different tasks from the ERASER benchmark~\cite{deyoung-etal-2020-eraser}. These include Movies for sentiment analysis~\cite{Pang2004ASE}, e-SNLI~\cite{NEURIPS2018_4c7a167b} for natural language inference, FEVER~\cite{thorne-etal-2018-fever} for fact extraction and verification, BoolQ~\cite{clark-etal-2019-boolq} for reading comprehension, and Evidence Inference~\cite{lehman-etal-2019-inferring} over scientific articles for medical interventions. All datasets except e-SNLI contain text spans or sentence level annotations as rationale. 
Following prior works~\cite{deyoung-etal-2020-eraser, pruthi2020evaluating, paranjape-etal-2020-information} we report F1 measures for both the task and token-level rationale extraction performance.

\begin{table}[h]
\small
\begin{tabular}{c|c|c|c|c}
\toprule
Dataset & \#Class & \#Full Train & \#Validation & \#Test \\
\midrule
Movies & 2 & 1.6K & 200 & 200 \\
e-SNLI & 3 & 547K & 9.8K & 9.8K  \\
FEVER & 2 & 98K & 6.1K & 6.1K \\
BoolQ & 2 & 6.3K & 1.4K & 2.8K \\
Evidence & 3 & 7.9K & 972 & 972 \\
\bottomrule
\end{tabular}
\vspace{-1em}
\caption{ Dataset statistics.}
\vspace{-0.5em}
\end{table}

\begin{table*}[ht]
\centering
\resizebox{\textwidth}{!}{%
\begin{tabular}{lcccccccccccc}
\toprule
\textbf{Model}
& \multicolumn{2}{c}{\textbf{Movies}}
& \multicolumn{2}{c}{\textbf{ESNLI}}
& \multicolumn{2}{c}{\textbf{FEVER}}
& \multicolumn{2}{c}{\textbf{BoolQ}}
& \multicolumn{2}{c}{\textbf{Evidence}} 
& \multicolumn{1}{c}{\textbf{Avg.}}\\
& \small{Task.} & \small{Tok.F1}
& \small{Task.} & \small{Tok.F1}
& \small{Task.} & \small{Tok.F1}
& \small{Task.} & \small{Tok.F1}
& \small{Task.} & \small{Tok.F1}
& \small{Task.} \\
\midrule
\textbf{\emph{fully supervised}}  & & & & & & & \\
\,BERT w/o explanation & 91 & -- & 90 & -- & \textbf{91} & -- & 62 & -- & 53 & -- & 77 \\ 
\,BERT with explanation  & \textbf{93} & 41  & 90 & 30 & \textbf{91} & 80 & \textbf{65} & 50 & 54 & 51 & \textbf{79} \\ 
\,BERT-to-BERT & 87 & 15 & 73 & 70 & 88 & 81 & 54 & 13 & \textbf{71} & 47  & 75 \\ 
\,\citet{pruthi2020evaluating} & 84 & -- & -- & -- & -- & -- & -- & -- & -- & -- & --\\ 
\citet{paranjape-etal-2020-information} (25\% rationales) & $85$ & 28 & -- & -- & 89 & 64 & 63 & 19 & 46 & 10 & 71 \\ 
\midrule
\textbf{\emph{few-shot}} \emph{($100$ labels per class)}\\
\,BERT w/o explanation & 82 & -- & 59 & -- & 69 & -- & 52 & -- & 41 & -- & $61$ \\ 
\,BERT with explanation & $82$ & 18 & $61$ & 5 & $70$ & 15 & $56$ & 16 & 41 & 30 & $62$ \\ 
\,Our model & \textbf{86} & 21 & \textbf{67} & 15 & \textbf{74} & 51 & \textbf{61} & 18 & \textbf{43} & 19 & \textbf{66} \\ 
\midrule
\textbf{\emph{few-shot}} \emph{(25\% of training data)} \\
\,BERT w/o explanation & $83$ & -- & 86 & -- & 88 &  & 59 & -- & 43 & -- & 72 \\
\,BERT with explanation & $82$ & 26 & 86 & 27 & \textbf{89} & 32 & $60$ & 31 & 44 & 8 & 72 \\
\,Our model & \textbf{87} & 31 & \textbf{87} & 36 & \textbf{89} & 39 & \textbf{62} & $51$ & \textbf{46} & 9 & \textbf{74} \\
\bottomrule
\end{tabular}
}
 \vspace{-0.5em} \caption{Comparison of task F1 (Task.) and rationale token F1 (Tok. F1) of our model with baselines. Avg. denotes the aggregated task performance averaged across all tasks. Our model includes BERT with explanation and rationale characteristics with self-training Baseline numbers are reported from corresponding papers.
}
 \vspace{-1em}
\label{tab:table results}
\end{table*}

\paragraph{\bf Base encoder} Following prior work~\cite{deyoung-etal-2020-eraser}, we adopt BERT-base~\cite{devlin-etal-2019-bert} as the base encoder for our experiments. We use a maximum sequence length of $512$ with a batch size of $8$. We use the Adam~\cite{kingma2017adam} optimizer with a learning rate of $3e-5$. Following~\citep{paranjape-etal-2020-information}, we use domain based checkpoint (BioBert\footnote{https://huggingface.co/bionlp/bluebert\_pubmed\_uncased\_L\-12\_H\-768\_A\-12}) for Evidence Inference. For Movies, we use a BERT checkpoint pre-trained on IMDB reviews. For all other datasets, we initialize the models from publicly available checkpoints\footnote{https://huggingface.co/bert-base-uncased}. Due to the question-answering format of tasks in BoolQ, Evidence Inference and FEVER, the query or claim with supporting documents are encoded in the form \texttt{document [SEP] query}.

\paragraph{\bf Few labels setup} We consider a resource-constrained setting with only $N = 100$ labeled samples per class and their associated rationales for each task, which are randomly sampled from the training data. We use the rest of the examples in the training set as unlabeled examples by ignoring their labels as in prior work~\cite{NEURIPS2020_f23d125d}. 
We perform early stopping on validation performance for the teacher multi-task training in each loop of self-training. We select the best model based on validation loss from the different self-training iterations. 
For Evidence, we use a hyper-parameter\footnote{Set as inversely proportional to the count of pseudo-labeled samples per class} to upweight examples from the minority pseudo-labeled class in the self-training loop to combat class imbalance.

\paragraph{\bf Baselines} We compare our model performance to the following methods for both \textit{fully supervised} (with access to all training labels) and our \textit{few-shot} setting with $100$ labels per class. (1) \textit{BERT w/o explanation} fine-tunes BERT on a set of labeled examples without accounting for rationales. (2) \textit{BERT with explanation} is our multi-task learning setup where the classifiers are trained to predict both task labels and rationales {\em without} encoding the rationale characteristics. 
We also compare against the \textit{semi-supervised} setting from prior work~\citet{paranjape-etal-2020-information} that uses 25\% annotations for rationales and 100\% task labels for each dataset.

Finally, we compare our method against a fully supervised BERT-to-BERT model in the ERASER benchmark~\cite{deyoung-etal-2020-eraser} which first performs rationale extraction followed by downstream task prediction using extracted rationale.

\begin{figure*}
    \centering
    \subfigure[]{\includegraphics[width=0.24\textwidth]{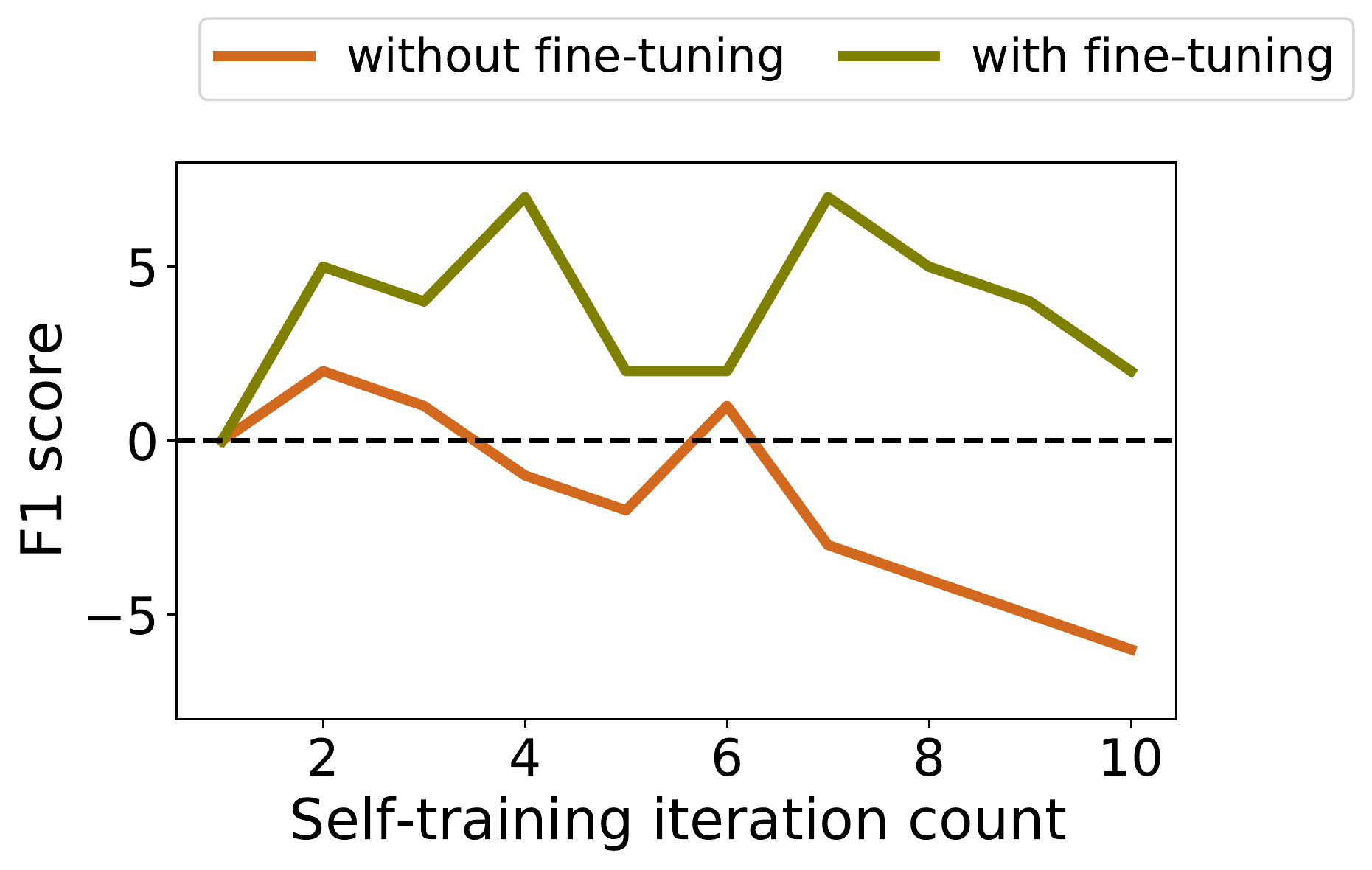}} 
    \subfigure[]{\includegraphics[width=0.24\textwidth]{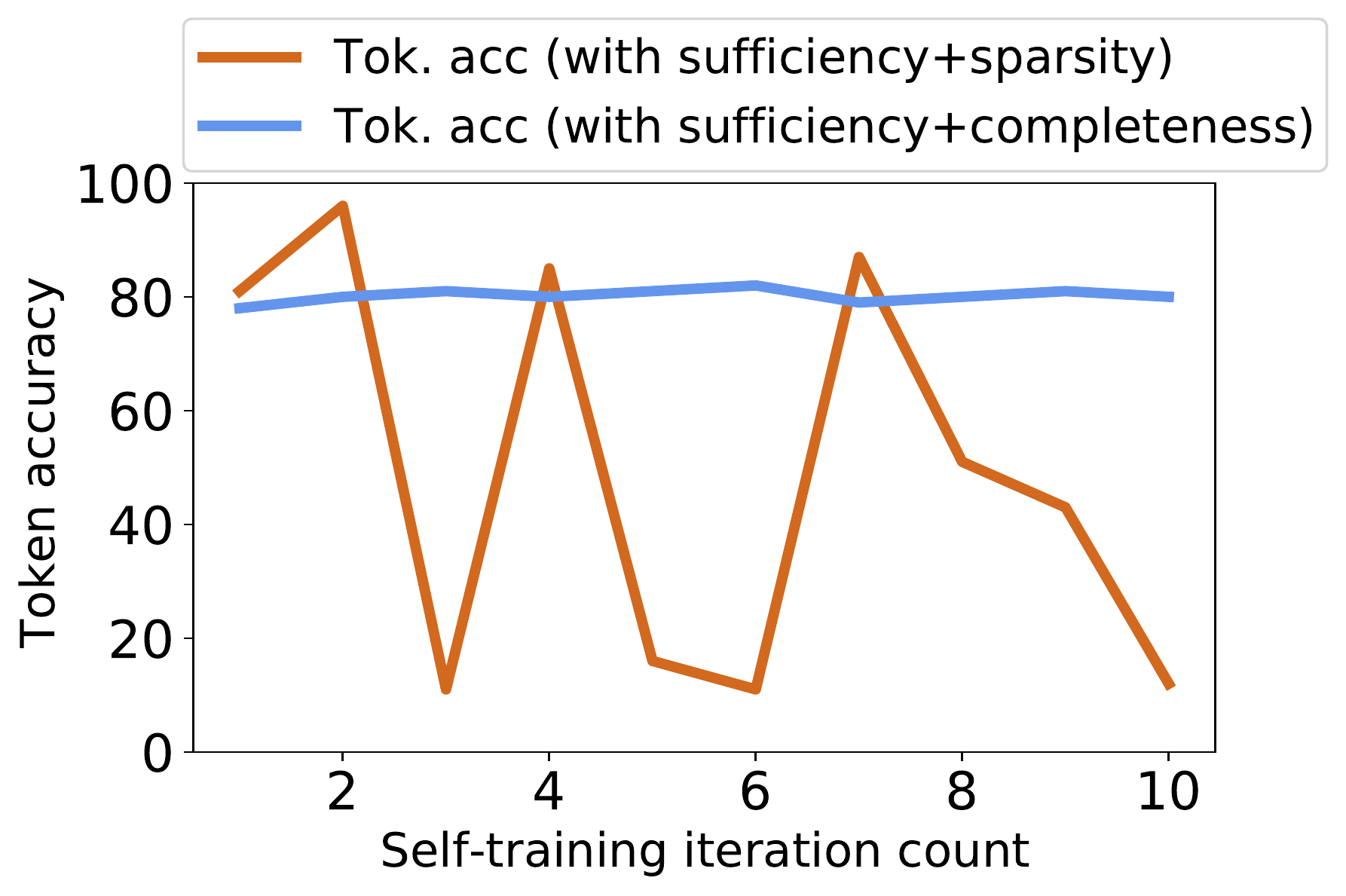}}
    \subfigure[]{\includegraphics[width=0.24\textwidth]{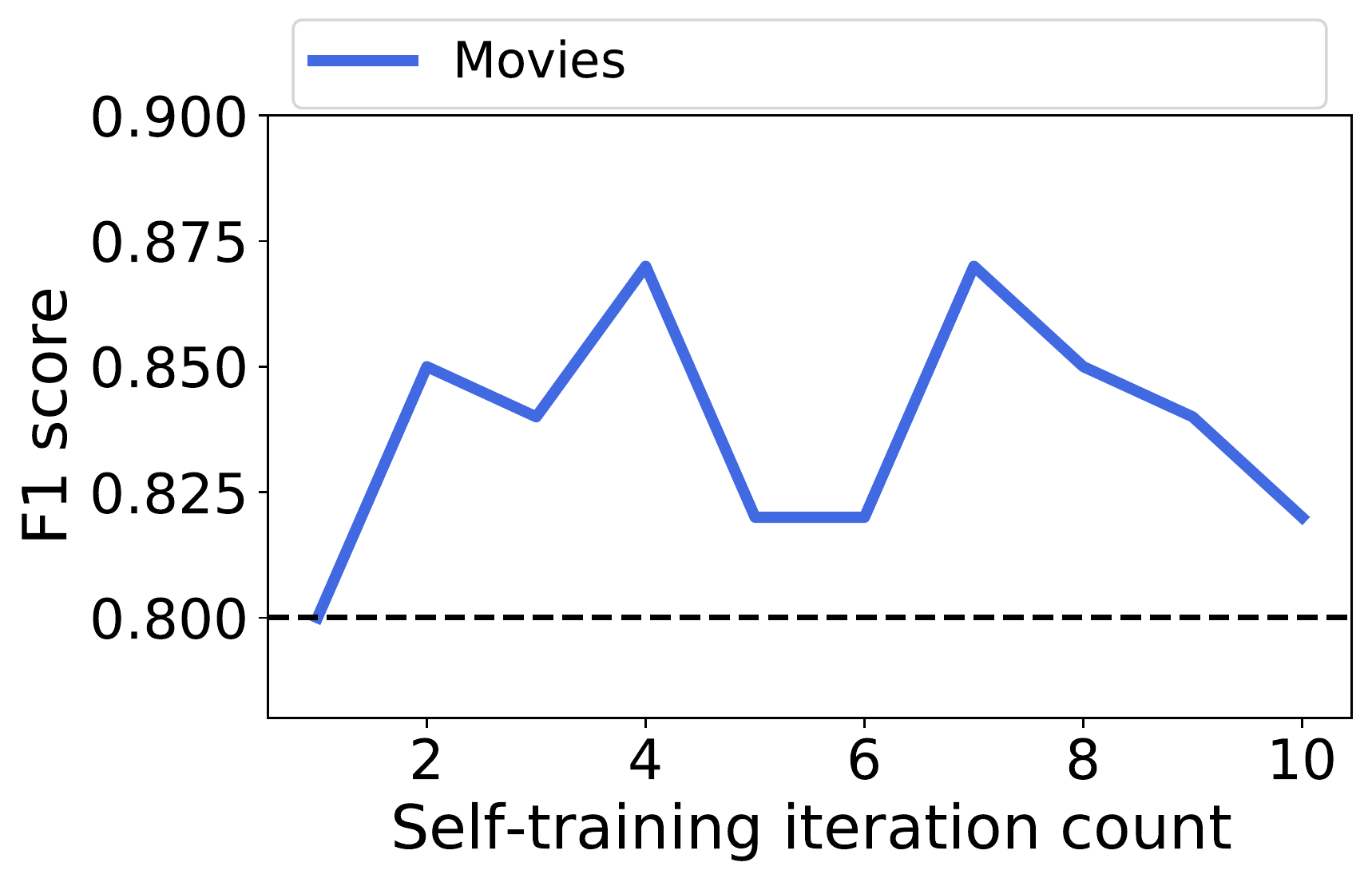}}
    \subfigure[]{\includegraphics[width=0.24\textwidth]{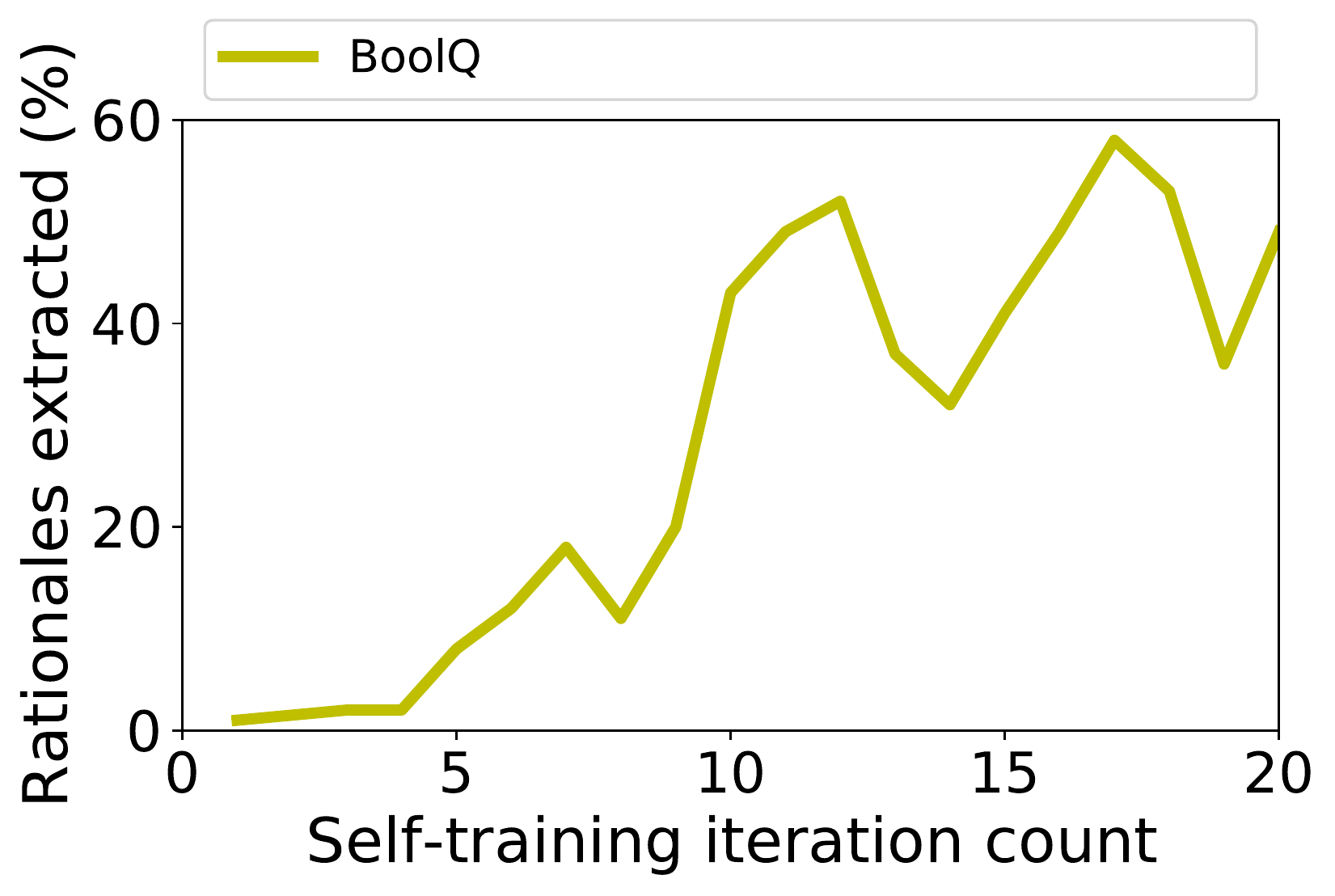}}
    \caption{Graphs representing model performance from different ablation study experiments: (a) Impact of fine-tuning with labeled data in self-training loop (b) Ablation study showing impact of adding sparsity loss without completeness, where we observe inconsistent rationale extraction in self-training. (c) Accuracy of Movies dataset across self-training iterations (d) Percentage of rationales extracted for BoolQ.}
    \label{fig:ablation}
\end{figure*}

\begin{table*}[h]
\centering
\footnotesize
\begin{tabular}{p{5cm}|ccc|ccc}
\toprule
\textbf{Model} & \multicolumn{3}{c}{\textbf{Movies}} & \multicolumn{3}{c}{\textbf{e-SNLI}} \\
 & \small{\textit{BLEU-2}} & \multicolumn{1}{c}{\small{\textit{Token P/R/F1}}}  & \small{\textit{Task.}} & \small{\textit{BLEU-2}} & \multicolumn{1}{c}{\small{\textit{Token P/R/F1}}} & \small{\textit{Task.}}\\
\midrule
teacher & 13.0  & 24.1/ 50.1/ 32.5 & 67 &  0  & 0.7/ 5.2/ 1.4 & 56 \\
\midrule
sufficiency & 10.1  & 22.1/ 93.1/ 35.7 & 69 & 0.98  & 10.0/ 12.1/ 10.0 & 61 \\
\midrule
sufficiency + re-weight labels & 9.3  & 9.1/ 96.4/ 16.6 & 70 & 2.0 & 34.4/ 14.6/ 20.5 & 65  \\
\midrule
sufficiency + re-weight labels and rat. & 12.2  & 19.6/ 92.2/ 32.3 & 71 & 1.3 & 33.1/ 10.7/ 16.2 & 65 \\
\midrule
sufficiency + sparsity + re-weight labels and rat. & 14.1  & 19.8/ 61.3/ 29.9 & 70 &  1.1  & 19.3/ 15.7/ 17.3 & 65 \\
\midrule
sufficiency + sparsity + completeness + re-weight labels and rat. & 14.0  & 24.2/ 50.3/ 32.3 & 71 & 1.07 & 12.8/ 7.6/ 5.7 & 65 \\
\midrule
Our model - re-weight labels and rat. & 12.7 &  21.2/ 47.6/ 29.3 & 71 &  1.6  & 15.6/ 11.4/ 13.17 & 65 \\
\midrule
Our model - coherence loss & 13.1 &  25.6/ 44.1/ 31.8 & 73 &  6.1  & 16.4/ 23/ 18.3 & 66 \\
\midrule
Our model & 14.1  & 28.4/ 41.3/ 33.5 & 75 &  5.0 & 17.4/ 13.8/ 15.3 & 67  \\
\bottomrule
\end{tabular}
\caption{Ablation study of various components in our model with corresponding performance in terms of BLEU-2, Token F1 and Task F1 with sequence length $128$. }
\label{tab:ablation results}
\end{table*}
\section{Results}
\paragraph{\bf Overall performance} Table~\ref{tab:table results} summarizes the results of our proposed model and baselines across all the datasets. We observe that our model trained with only $N=100$ labels per class performs within $10.6\%$ of fully supervised BERT trained with thousands of labels while obtaining an aggregate F1 of $66\%$. Our self-training framework iteratively improves over the teacher model with a judiciously designed student network with a performance gain of $6.45\%$.

For the fully supervised models, we observe that BERT with explanation in our multi-task learning setup improves by $2.6\%$ over vanilla BERT, thereby, demonstrating the usefulness of rationales and the effectiveness of multi-task learning.

Our model performance for downstream tasks and rationales further improves with increasing the amount of labeled data. With $25\%$ labeled training data, our model performs at par with \citet{paranjape-etal-2020-information} (which has been trained with 100\% task labels) in $3$ out of $4$ tasks and has comparable performance with fully supervised BERT-to-BERT baseline.
Additionally, our model does not require additional user inputs in the form of desired sparsity threshold as in prior work~\citep{paranjape-etal-2020-information}. 

In order to validate the effectiveness of sufficiency and completeness loss as a way of integrating rationales in the model in contrast to attention regularization in~\citet{pruthi2020evaluating}, we perform the following experiment. Instead of using teacher-predicted rationales, we use the ground-truth rationales. We observe that our model with sufficiency and completeness loss outperforms the prior method of integrating explanations via attention regularization (results in Figure ~\ref{fig:pruthi-comparison} in Appendix). This may reflect the fact that attention regularization imposes the stricter assumption to have uniform attention on the explanation tokens.


We observe the self-training performance to improve on initializing our encoders with pre-trained domain-specific checkpoints. For instance, using BioBERT checkpoint, we observe a gain of $4.8\%$ over that of BERT-base in the Evidence dataset. 
Table~\ref{tab:good_examples} presents few examples from our rationale extractor and the corresponding task label.

\begin{table*}[h]
\footnotesize
\begin{tabular}{p{\textwidth}}
\toprule
\textbf{Dataset}: Movies \qquad \textbf{Ground truth}: Negative, \textbf{Prediction}: Negative \\
\midrule
\colorbox{lime}{There're so many things to criticize about I don't know where to start. Recommendation: turn off your brain} \colorbox{pink}{- don't be like me, } \colorbox{lime}{decreasing the rating everyday because I think about it too much}..... Firstly, there is \colorbox{yellow}{nothing outstandingly inferior}
\colorbox{yellow}{about the making of the film (nor is there anything outstandingly good about it),} but the plot holes make the film corny and stupid. \\
\midrule
\textbf{Dataset}: Movies \qquad \textbf{Ground truth}: Negative, \textbf{Prediction}: Negative \\
\midrule
\colorbox{lime}{Yet another brainless teen flick,} ......
\colorbox{yellow}{stars Katie Holmes and Sarah Polly couldn't look more bored.} One thing you need \\
to know is \colorbox{lime}{I really hated this movie. Everything about it annoyed the hell out of me.
The acting, and script, the plot, and ending.} \\
\midrule
\textbf{Dataset}: e-SNLI \qquad \textbf{Ground truth}: Contradiction, \textbf{Prediction}: Contradiction \\
\midrule
A man \colorbox{pink}{playing} \colorbox{lime}{electric guitar} on the \colorbox{lime}{stage}. A man playing \colorbox{lime}{banjo} on the \colorbox{lime}{floor}. \\
\bottomrule
\end{tabular}
\caption{Snapshot of correctly classified examples and the corresponding rationales extracted by our model on Movies and e-SNLI.\\ }
\label{tab:good_examples}
\begin{tabular}{p{\textwidth}}
\toprule
\textbf{Dataset}: Movies \qquad \textbf{Ground truth}: Positive, \textbf{Prediction}: Negative \\
\midrule
Well I'll be damned ... \colorbox{lime}{the Canadians can make a good movie.} The world is coming to an end. We don't know why or how , \\ \colorbox{pink}{but apparently there is no way to stop it}..... \colorbox{pink}{most of the rioting and other assorted chaos} has passed.  \colorbox{yellow}{director - star Don Mckellar } \colorbox{yellow}{has crafted a highly unique and emotional film.} \colorbox{yellow}{All of the main characters are compelling} as they try and do whatever it is..... \\
\midrule
\textbf{Dataset}: Movies \qquad \textbf{Ground truth}: Positive, \textbf{Prediction}: Negative \\
\midrule 
When I first saw the previews for Ron Howard's latest film, \colorbox{pink}{my expectations were discouragingly low.} \colorbox{pink}{A show about nothing?} A guy whose entire life is broadcast 24 hours a day ?....  which is why \colorbox{yellow}{I was pleasantly surprised by "edtv,"} \colorbox{yellow}{which turns out to be a fresh , insightful , and often times hilarious film about the follies of instant celebrity}.\\
\midrule
\textbf{Dataset}: e-SNLI \qquad \textbf{Ground truth}: Entailment, \textbf{Prediction}: Neutral \\
\midrule 
A \colorbox{yellow}{woman} tired from her long day takes a \colorbox{lime}{nap} on her bed above the sheets and covers. A \colorbox{lime}{lady} is \colorbox{yellow}{lying in bed.} \\
\bottomrule
\end{tabular}
\caption{Snapshot of mis-classified examples from our model on Movies and e-SNLI dataset. 
\newline
\textit{Legend:} \colorbox{yellow}{Ground-truth rationales not detected by the model} \colorbox{lime}{Rationales present in both the model and ground-truth} \colorbox{pink}{Rationales extracted by the model but absent in ground-truth}
}
\label{tab:examples}
\end{table*}


\subsection{Ablation Study}
Table~\ref{tab:ablation results} summarizes the impact of different components of our framework with $N=100$ labels per class as training data for Movies and e-SNLI.

\paragraph{Re-weighting pseudo-labels} We found it extremely useful to re-weight noisy pseudo-labeled samples from the teacher model by its confidence. We observe that re-weighting the rationales and task labels work quite well for the Movies dataset. However, this has limited impact for e-SNLI with a low coverage of rationales from the teacher model. Correspondingly, we did not observe a difference in model performance from re-weighting the task and rationale pseudo-labels. 

\paragraph{\bf Impact of different loss functions.} We observe that using only the sufficiency loss (Eq.~\ref{eqn:suff_loss}) results in the model extracting the entire input as the rationale. This is counteracted by adding penalization via sparsity loss to obtain rationales that are concise yet informative about the task label. From Table~\ref{tab:ablation results}, we observe sparsity loss to significantly reduce the number of tokens included in the rationale. 

However, adding the sparsity loss also caused instability in some cases (Figure~\ref{fig:ablation} (b)).
We empirically demonstrate that this instability is mitigated by including the completeness loss that forces the model to be maximally uncertain when it does not look at important tokens constituting the rationale. 


\paragraph{\bf Impact of amount of labeled training data} We observe our self-training framework to improve both in task and rationale extraction performance with increase in number of labeled samples for training (Appendix, Figure~\ref{fig:varying_N}). 

\paragraph{\bf Impact of number of self-training iterations} Figure~\ref{fig:ablation} (c) shows the improvement in the task accuracy of our model over several self-training iterations for Movies. The corresponding plots for other datasets are provided in Figure~\ref{fig:accuracy_datasets} in Appendix. 
We observe that self-training gradually improves model performance in the first few iterations for majority of the tasks and converges fast in 12-15 iterations. However, we observe that rationale extraction module drifts after $10$ self-training iterations for most of the datasets (Figure~\ref{fig:ablation} (d)) due to error propagation from noisy pseudo-labels, thereby, necessitating early stopping based on rationale validation loss. \\

\paragraph{\bf Few labeled data fine-tuning} At each self-training iteration, the teacher is fine-tuned on labeled data. This is to avoid drifting from the original task description via few annotated labels (Figure~\ref{fig:ablation} (a)). Figure~\ref{fig:ablation} demonstrates the change in the accuracy of our student model with and without this teacher training in every self-training iteration.

\subsection{Error Analysis} Table~\ref{tab:examples} presents a snapshot of the qualitative error analysis of our model. On analyzing the extracted rationales for mis-classified instances, we observe some common failure points where instances have shifts in context; presence of satire or sarcasm; rationales relying on background knowledge; and noisy or incomplete annotated rationales. 
For instance, the overall polarity of the first example from Movies is positive, although majority of the text describing the movie plot depicts a negative connotation.
We observe a similar trend with reviews involving {sarcasm or satire}. In the last example from e-SNLI, the annotators marked {\em \{woman, nap\}} to {\em entail} {\em \{lady, lying in bed\}}. In this rationale, annotators do not follow the guidelines for sufficiency and completeness since the spatial qualifier for {\em nap} is missing in the ground-truth. Surprisingly, our model does not pick up the spatial concept either and marks the sequences to be {\em neutral} to each other. 


\section{Conclusion}
We develop a multi-task self-training framework for rationale extraction focusing on low-resource settings with access to very few training labels. To this end, we build on insights from prior work on the characteristics of a good rationale to encode them via judiciously designed loss functions in our self-training framework. 
Extensive experiments on benchmark datasets show our model to outperform other state-of-the-art methods with access to limited labels. We further demonstrate that the performance of pre-trained language model can be improved by making it aware of the rationales for its decision-making process in both high (fully supervised) and low-resource (few label) settings.



\bibliography{anthology,custom}
\bibliographystyle{acl_natbib}

\newpage
\appendix
\section{Appendix}
\label{sec:appendix}

\begin{figure}[!htb]
\includegraphics[scale=0.45]{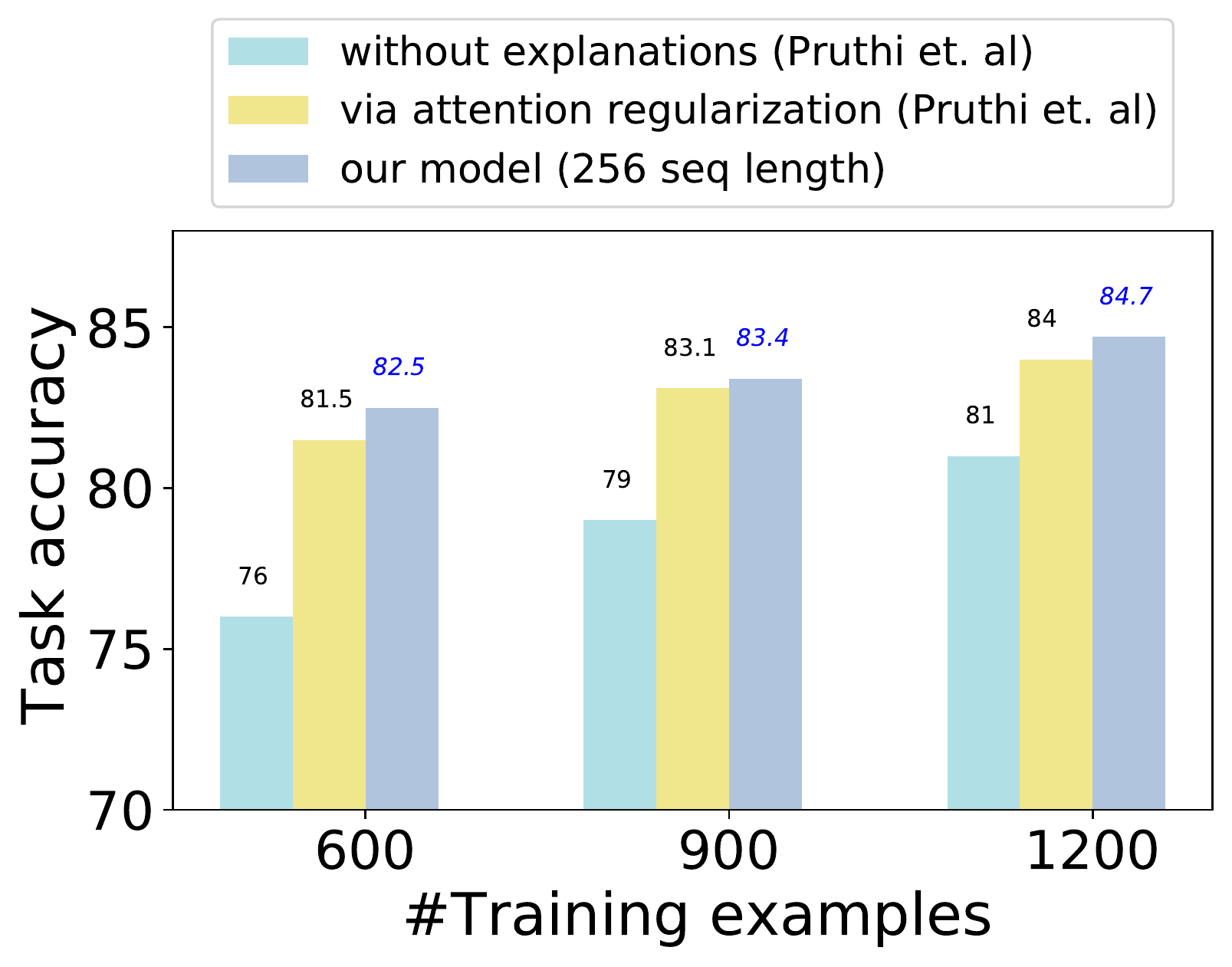}
\caption{Integrating rationales with sufficiency $\mathcal{L}_{suff}$ and completeness $\mathcal{L}_{comp}$ losses outperforms the recent attention regularization method~\citep{pruthi2020evaluating}. To be comparable, we use only labeled examples, we don't do any self-training (no unlabeled examples), and we use max sequence length of 256.}
\label{fig:pruthi-comparison}
\end{figure}

\begin{figure}[htb]
\includegraphics[scale=0.5]{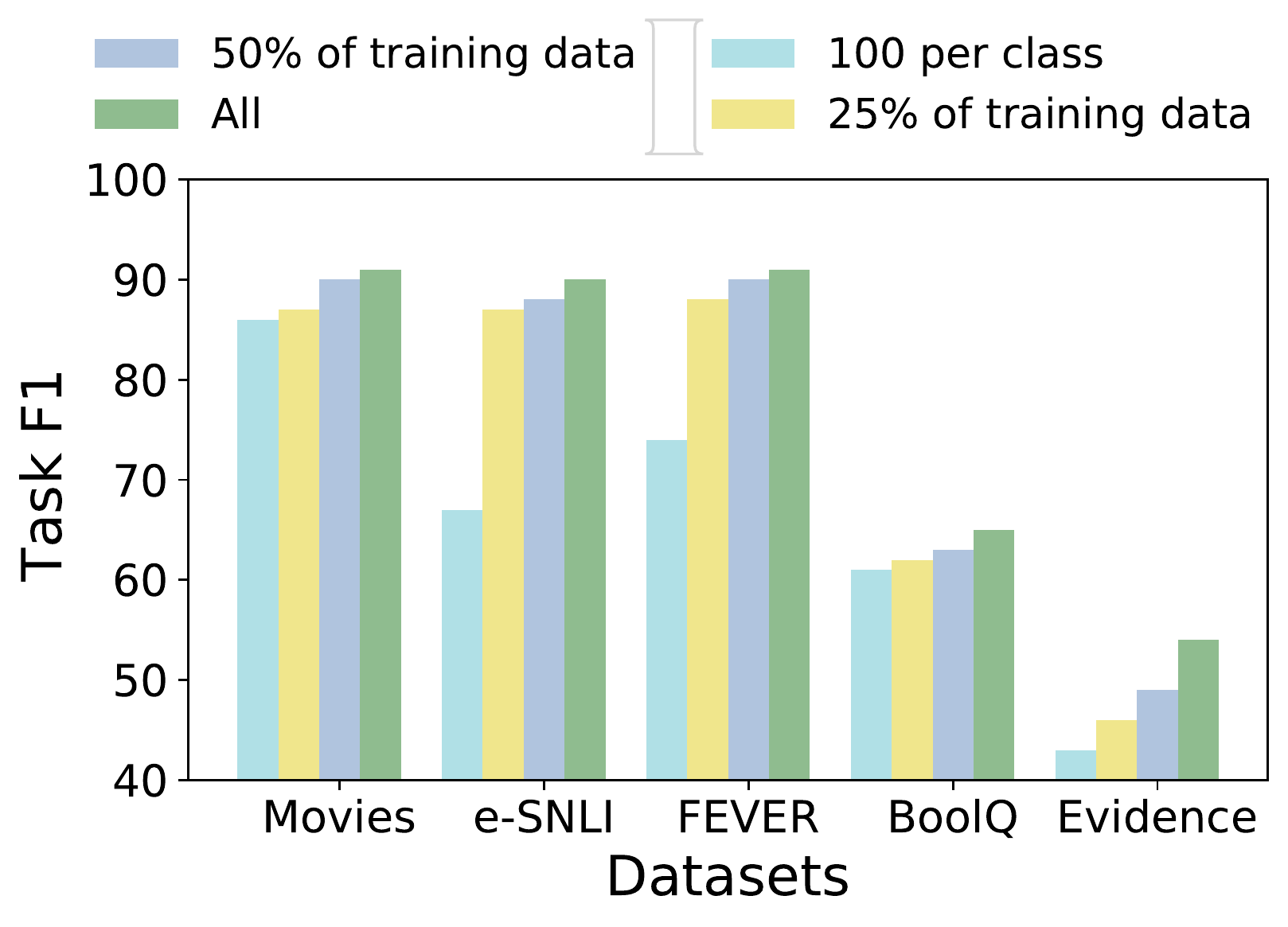}
\caption{Task F1 across varying percentage of training data across datasets}
\label{fig:varying_N}
\end{figure}
\clearpage

    \begin{figure*}[h!]
        \includegraphics[width=.50\textwidth]{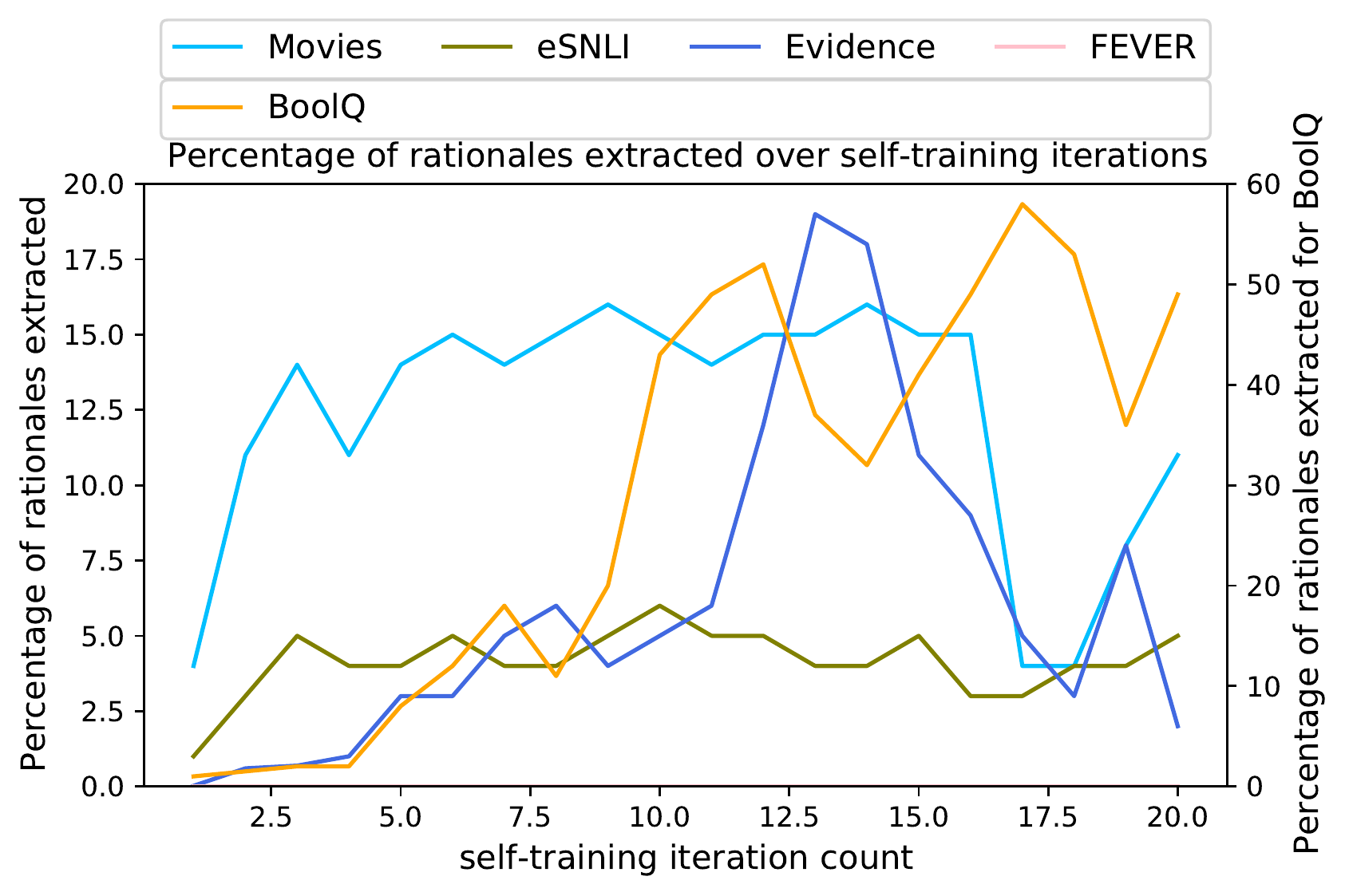}
    \includegraphics[width=.50\textwidth]{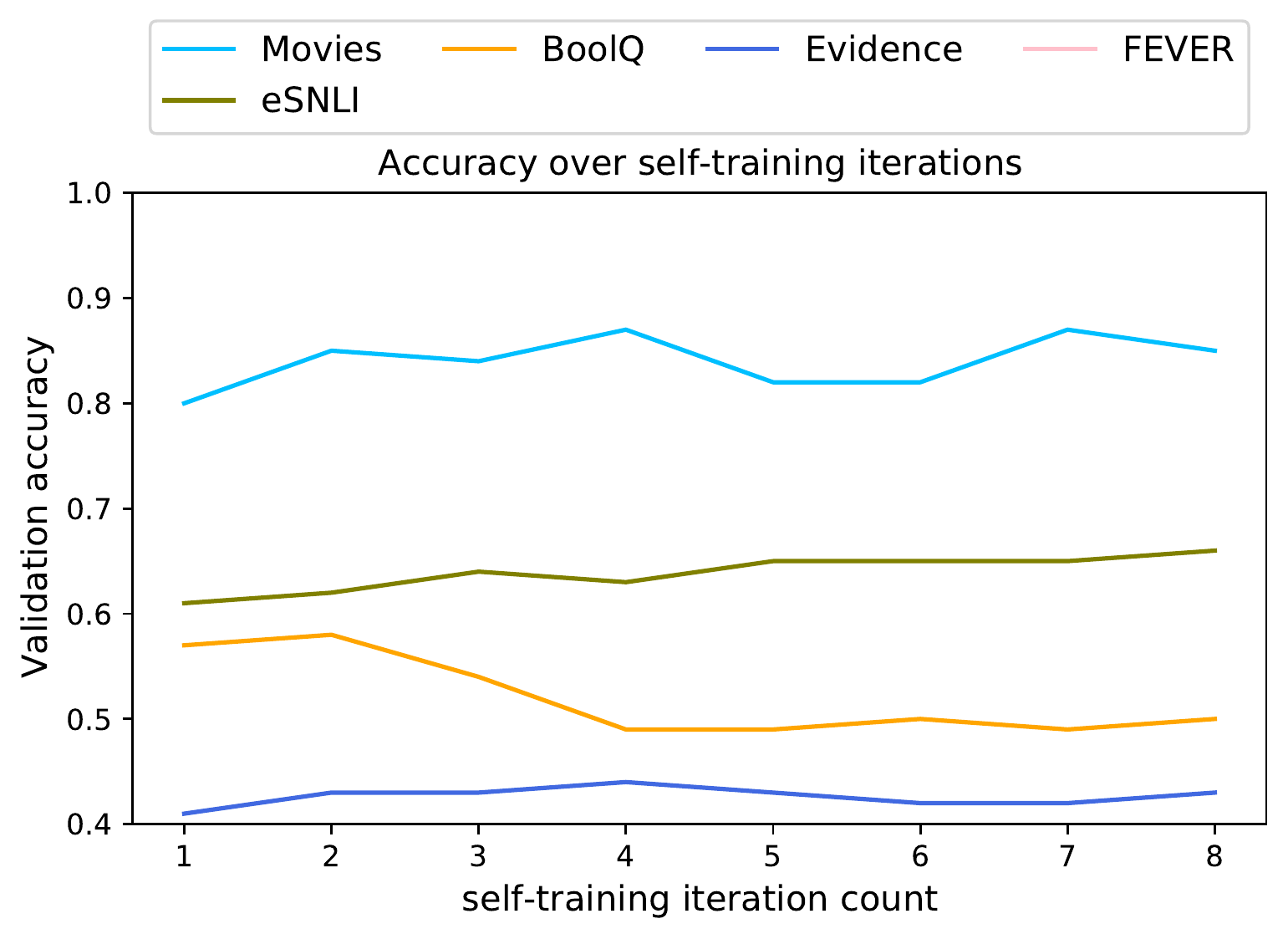}
    \caption{Percentage of rationales extracted over self-training iterations}
    \caption{Accuracy of all the datasets across self-training iterations.}
    \label{fig:per_rationales}
    \label{fig:accuracy_datasets}
\end{figure*}

\begin{table*}[h!]
\footnotesize
\begin{tabular}{l}
\hline
Dataset: Movies \qquad \textit{Ground truth: Negative}, \textit{Prediction: Positive} \\
\midrule
Though \colorbox{pink}{good - looking , its lavish sets , fancy costumes and luscious cinematography} can do \\ little to compensate for the emotional wasteland... this is Jodie Foster 's first \\ movie since \colorbox{pink}{the jaw - droppingly brilliant} contact came out more than two years ago \\ and \colorbox{yellow}{it isn't the best choice to show off her acting chops .} \\
\midrule
Dataset: e-SNLI \\
\midrule
\textit{Ground truth: Entailment}, \textit{Prediction: Neutral} \\
A \colorbox{yellow}{woman} tired from her long day takes a \colorbox{lime}{nap} on her bed above the sheets and covers. A \colorbox{lime}{lady} is \colorbox{yellow}{lying in bed.} \\
\midrule
\textit{Ground truth: Contradiction}, \textit{Prediction: Entailment} \\
A \colorbox{pink}{mountainous photo} is complete with a \colorbox{yellow}{blue sky.} The photo was taken on a \colorbox{lime}{cloudy night.} \\
\hline
\end{tabular}
\caption{Snapshot of mis-classified examples from Movies and ESNLI.
\\ \textit{Legend} \colorbox{yellow}{Ground-truth rationales not detected by the model}, \colorbox{pink}{Rationales extracted by the model but absent in ground-truth.} \colorbox{lime}{Rationales present in both the model and ground-truth.}}
\label{tab:examples_appendix}
\end{table*}

\end{document}